\documentclass[letterpaper, 10 pt, conference]{ieeeconf}  

\IEEEoverridecommandlockouts                              

\usepackage{amsmath} 
\usepackage{multirow}
\usepackage{makecell}
\usepackage{array}
\usepackage{graphicx}
\usepackage{stfloats}
\usepackage[
    singlelinecheck=false
]{caption}
\title{\LARGE \bf
Multi-modal anticipation of stochastic trajectories in a dynamic environment with Conditional Variational Autoencoders
}

\author{Albert Dulian$^1$ and John C. Murray$^2$
\thanks{The authors are with: 1 - Department of Computer Science and Technology, The University of Hull, Kingston Upon Hull, United Kingdom; 2 - University of Sunderland, Faculty of Technology, Sunderland, United Kingdom {\tt\small A.Dulian-2013@hull.ac.uk, John.Murray@sunderland.ac.uk}}}

\begin{document}
\maketitle
\thispagestyle{empty}
\pagestyle{empty}
\begin{abstract}

Forecasting short-term motion of nearby vehicles presents an inherently challenging issue as the space of their possible future movements is not strictly limited to a set of single trajectories. Recently proposed techniques that demonstrate plausible results concentrate primarily on forecasting a fixed number of deterministic predictions, or on classifying over a wide variety of trajectories that were previously generated using e.g. dynamic model. This paper focuses on addressing the uncertainty associated with the discussed task by utilising the stochastic nature of generative models in order to produce a diverse set of plausible paths with regards to tracked vehicles. More specifically, we propose to account for the multi-modality of the problem with use of Conditional Variational Autoencoder (C-VAE) conditioned on an agent's past motion as well as a rasterised scene context encoded with Capsule Network (CapsNet). In addition, we demonstrate advantages of employing the Minimum over N (MoN) cost function which measures the distance between ground truth and N generated samples and tries to minimise the loss with respect to the closest sample, effectively leading to more diverse predictions. We examine our network on a publicly available dataset against recent state-of-the-art methods and show that our approach outperforms these techniques in numerous scenarios whilst significantly reducing the number of trainable parameters as well as allowing to sample an arbitrary amount of diverse trajectories.

\end{abstract}

\begin{figure}[t]
  \centering
   \includegraphics[scale=.9]{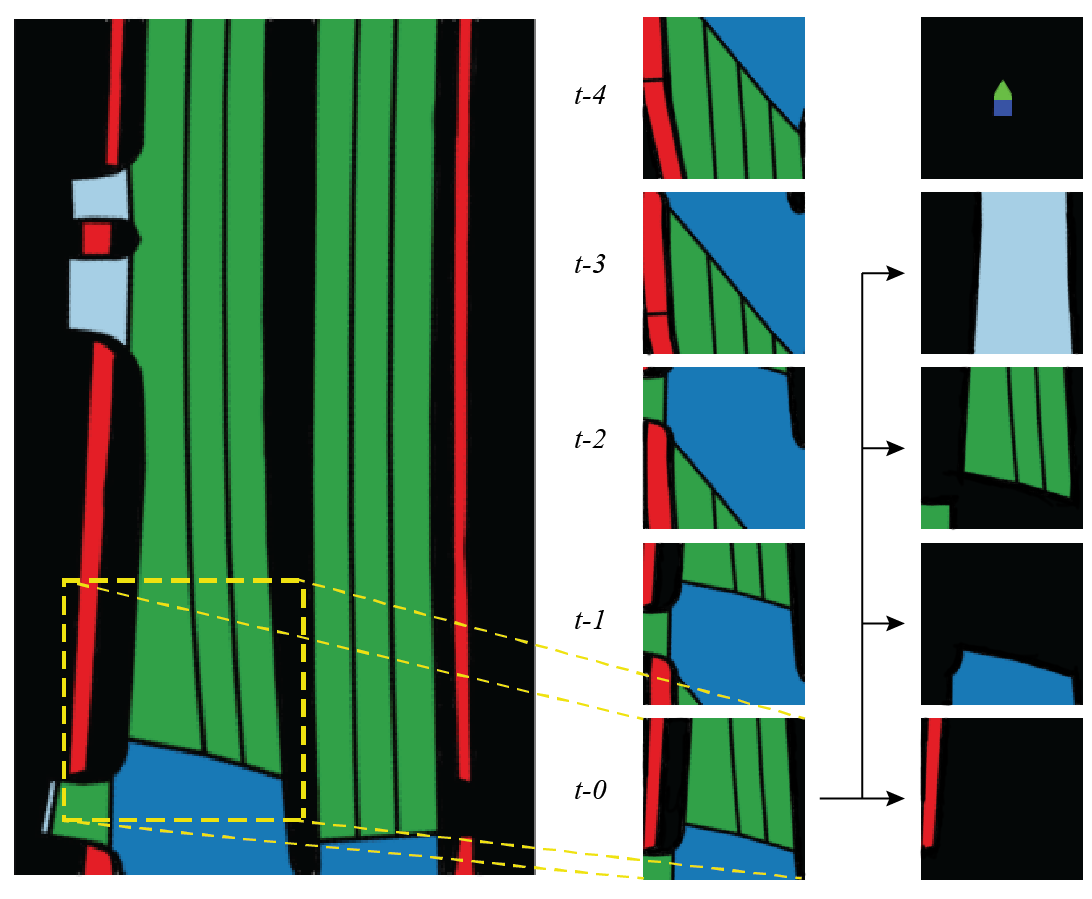}
  \caption{An example of a global map $\mbox{\boldmath$M$}$ (left) as well as the unraveling process of a local map chunk $\mbox{\boldmath$\mathsf L$}_{t}$ (middle) which is turned into several geometrical layers (right) each representing a single part of the whole map. In addition, we draw an agent (top right) with its origin corresponding to the position on the chunk of interest.}
  \label{fig:map_dis}
\end{figure}
\section{INTRODUCTION}
In recent years we have seen a dramatic increase in interest within the area of autonomous transportation and its associated research. Higher levels of autonomy have a great potential to offer robust and effective solutions that will have significant impact in a reduction of on-road risk. However, despite great developments in the intelligent technology that autonomous vehicles (AVs) are being equipped with there still exists a number of tasks that pose a variety of complicated challenges. For instance, it is critical for AVs to be equipped with capabilities that allow them to accurately track other agents and AVs, predicting their future states, such as their future trajectories, over a short period of time e.g. 4 seconds. This will enable, for instance, the ability to conduct further analysis with with respect to the detection of potentially dangerous on-road situations such as collisions.

This work aims to concentrate on predicting future paths of vehicles from a point of view of an autonomous vehicle navigating in a complex and uncertain environment. This problem can be further formulated as predicting or generating a sequence of future coordinates that describe the possible motion of a tracked agent over a specified time horizon. Yet an accurate anticipation of an agents' future movement is inherently difficult as it requires an output of a model to represent a distribution of plausible paths (multi-modal) as opposed to a single deterministic trajectory. We aim to tackle this challenging task by leveraging benefits of generative models, in particular a \textit{Conditional Variational Autoencoder} (C-VAE) \cite{sohn2015learning} that approximates a posterior distribution of future trajectories by conditioning the decoder on the past movement of a tracked agent, as well as on the global rasterised top-down image of the road. 

A number of recently suggested approaches have explored the use of various \textit{Deep Learning} (DL) \cite{goodfellow2016deep} methods to address the discussed task  \cite{messaoud2019non, gao2020vectornet, cui2019multimodal}, however, these still manifest several drawbacks such as being restricted to rather uncomplicated surroundings (e.g. highways), lack of multi-modality or in case of multi-modal prediction, deterministic outputs. On the other hand, over the past years VAEs \cite{kingma2013auto} as well as its variations (e.g. C-VAE) has shown a great promise in modelling of complex distributions for numerous tasks such as generating images \cite{gregor2015draw}, sentences \cite{bowman2015generating}, music \cite{roberts2017hierarchical}, as well as vehicle trajectories \cite{lee2017desire, choi2019drogon}. In addition, a majority of existing techniques utilise some form of High-Definition (HD) maps which encapsulate various road attributes such as lanes \cite{srikanth2019infer, cui2019multimodal, phan2020covernet, marchetti2020mantra}. The data from HD map is then often transformed (rasterised) into its top-down image representation and fed into a \textit{Convolutional Neural Network} (CNN) \cite{le1989handwritten} to encode the global scene context and extract salient features. Nonetheless, despite achieving state-of-the-art performance in several computer vision tasks \cite{tao2020hierarchical, touvron2020fixing} CNNs still exhibit two crucial drawbacks i.e. equivariance as well as local invariance \cite{goodfellow2016deep}. CNNs can only achieve equivariance with regards to translation (parameter sharing), however, applying other transformations, for instance rotation will often cause non-activation of relevant neurons within the network (unless data augmented through such transformation was present during training), thus failing to detect relevant features. In contrast, the issue of invariance has been often mitigated through the use of pooling operation, e.g. max-pooling, which computes a maximum value from sub-regions of an initial input, effectively leading to achieving small local invariance as well as reduction of dimensionality which in many cases leads to a loss of large volumes of valuable spatial data. Hinton \textit{et al.} \cite{hinton2011transforming} proposed to tackle these issues by introducing novel type of neural network called \textit{Capsule Network} (CapsNet) which uses capsules (locally invariant group of neurons in a vectorised form) that learn numerous properties of objects (e.g. their pose) within a spatial domain. Encoded parameters can be further conceptualised as an object’s instantiation parameters that enable the model to learn more robust and equivariant representation of features with respect to change in viewpoint. Lastly, we perform analysis of employing \textit{MoN} (Minimum over N) loss function (often called Variety Loss \cite{gupta2018social}) which has previously proven the encouragement of diverse sample generation during training. To summarise our contribution are:
\begin{itemize}
    \item We demonstrate the use of Capsule Networks that encodes the context of the environment from its global and local rasterised top-down image with respect the task of short-term motion prediction. 
    
    \item We propose a novel combination of CVAE and CapsNet trained with MoN loss which learns to produce a diverse set of stochastic trajectories in complex environments.
    
    \item We examine MoN on various settings with respect to 1. number of trajectories sampled during training stage and 2. different choice of distance functions that measure error between generated samples and ground truth motion.
    
    \item We examine and compare the performance of our proposed method against recent state-of-the-art methods on publicly available dataset.
\end{itemize}

\section{RELATED WORK}\label{sec:related}
The problem of forecasting short-term motion of vehicles has been widely studied in recent years. Traditionally, this matter has been approached with an assumption that the agent's future states can be well approximated using a physics based model e.g. constant velocity, constant turn rate and acceleration etc. A comprehensive review of such models can be found in \cite{schubert2008comparison}. Furthermore, collecting data through numerous sensors can introduce noise that can greatly reduce robustness as well as prediction accuracy of these methods. As a result, filtering algorithms such as \textit{Kalman Filter} (KF) \cite{kalman1960new} are often employed to minimise the noise in measurements \cite{ammoun2009real}. Regardless, such techniques can only achieve accurate estimation for a very short prediction horizon e.g. 1 second, and only if several parameters remain constant during computation time.

Classical physics based techniques impose unrealistic constraints that do not precisely reflect real world scenarios. \textit{Monte Carlo} methods \cite{hammersley2013monte} on the other hand provide an alternative solution in which the desired quantity can be approximated through random sampling. For instance, the work described in \cite{broadhurst2005monte} defines a framework based on Monte Carlo sampling to estimate a probability distribution of agent's future motion and then use sampled estimates to evaluate likelihood of future collisions. Moreover, probabilistic techniques such as \textit{Bayesian Networks} (BNs) \cite{jensen1996introduction}, \textit{Hidden Markov Models} (HMM) \cite{rabiner1989tutorial} and \textit{Gaussian Processes} (GP) \cite{rasmussen2003gaussian} has also been extensively examined with respect to the trajectory forecasting. In the following \cite{joseph2011bayesian} Joseph \textit{et al.} used mixture of GPs to obtain an adaptive representation of motion patterns derived from trajectory data where each pattern was quantified by a combination of location derivatives.

\begin{figure*}[t]
  \centering
  \includegraphics[scale=.2]{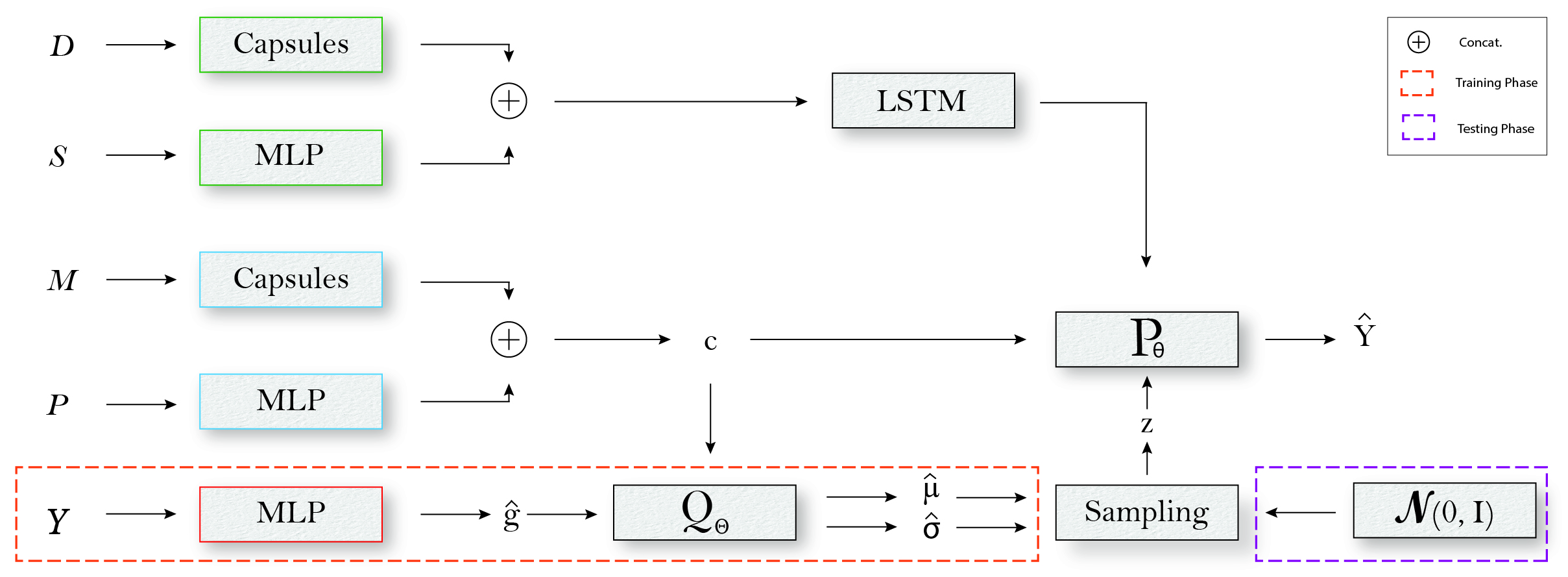}
  \caption{A simplified overview of the proposed network. During training we obtain the parameters of the prior distribution with recognition network which is used to produce a diverse set of training samples. Then, during testing the past ground-truth motion \mbox{\boldmath$\ Y$} of a tracked agent is not available and we therefore sample $z$ directly from prior i.e. $z \sim \mathcal{N}(\mbox{\boldmath$0$, \mbox{\boldmath$\mathbf{I} $}})$ and then decode $k$ samples which are encapsulated in \mbox{\boldmath$\hat{\mathsf Y}$}.}
  \label{fig:encoder_arch}
\end{figure*}

Beyond the traditional data driven methods which rely on hand-crafted features, there has been a great focus on employing DL based techniques in order to model complex dependencies between advanced datasets. For instance, Chandra \textit{et al.} introduced \textit{TraPHic} \cite{chandra2019traphic}, a model based on combination of LSTM \cite{hochreiter1997long} and CNN that addresses the issue of motion forecasting in dense traffic whilst simultaneously accounting for interactions between numerous heterogeneous on-road participants (e.g. bikes, buses). TraPHic captures relationships between traffic agents by defining a fixed horizon of interest (semi-elliptical region) using empirically determined radius as opposed to the conventionally employed grid-based methodology. The states of observed agents are then processed through a LSTM layer to account for the temporal dependencies, and further encoded through a CNN layer to learn local on-road interactions. As previously mentioned, a number of recent approaches utilised HD map data to improve the model's performance. For instance, the work from \cite{srikanth2019infer} constructs an intermediate representation of feature space from various bird's eye view map obtained from either stereo camera or LiDAR. Multitude of generated maps form a diverse semantic representations of the scene such as lanes, roads, obstacles etc, which are fed through a CNN-LSTM based network to predict location of the vehicle of interest on an occupancy grid map. Next, in the following \cite{lee2019joint} Lee \textit{et al.} introduced a novel approach based on \textit{Graph Neural Networks} (GNNs) \cite{battaglia2018relational} to model joint interactions between a pair of agents and reason about their future motion. Authors assumed that there exists a set of discrete interactions between pair of agents e.g. yielding to another driver, that can be learned from labeled data, thus allowing the GNN based model whose nodes and edges represented agents of interest as well as their long-term intents to learn those in a supervised fashion. Gao \textit{et al.} introduced \textit{VectorNet} \cite{gao2020vectornet}, an another example of uni-modal motion prediction model based on GNNs. However, as opposed to using rasterised HD maps to extract road features such as lanes, crosswalks etc, VectorNet transforms them into a vectorised representation that is further encoded with GNN. Comparison results between VectorNet and various pre-trained CNN architectures (e.g. ResNet-18 \cite{he2016deep}) indicate significant cost reduction of over 70\% whilst simultaneously increasing performance of the model on a short-term prediction (3 seconds). The work presented in \cite{dulian2021exploiting} demonstrates the use of CapsNet to encode local semantic chunks of the HD map for improved motion forecasting, however, it does not account for a multi-modal nature of the task and focuses only on a deterministic prediction. Next, Messaoud \textit{et al.} \cite{messaoud2020trajectory} stresses the importance of inferring trajectory cues from both scene structure and past motion data of interacting agents and proposed a model based on multi-head attention \cite{vaswani2017attention}. The discussed work suggests to encode and concatenate a joint representation of all agents of interests as well as a top-down view of static scene, and then divide it into a spatial grid to model future trajectory distribution as a mixture model. Furthermore, the following \cite{marchetti2020mantra} demonstrated an interesting approach by utilising the memory augmented network \cite{weston2014memory} to learn the association between past and future motion, and then memories most meaningful samples. In addition, the model manifested capabilities to constantly update its internal state through an online collection of observed samples. Lee \textit{et al.} also proposed to use CVAE conditioned on agents' past motion \cite{lee2017desire} by combining it with \textit{Inverse Optimal Control} (IOC) \cite{abbeel2004apprenticeship} based framework for assessing quality of predicted hypothetical trajectories. More recent use of CVAE was further demonstrated in \cite{choi2019drogon} with introduction of \textit{DROGON}, a model for multi-modal trajectory prediction. The proposed framework was trained by conditioning the stochastic model on the agent's estimated intentions as well as interactions with other traffic participants that were tracked with use of LiDAR data. Then, the input as well as sampled latent variable was decoded to predict number of heatmaps that encoded future locations as well as their likelihoods. In \cite{djuric2020uncertainty} Djuric \textit{et al.} argued that in order to learn essential features of the environment, one must transform its context into a meaningful representation, such as its rasterised top-down view, that will contain information with respect to road layout. The transformed context of the surrounding was used as input to CNN to learn salient features of the road to increase accuracy of predicted paths. Further extension of this work was demonstrated in \cite{cui2019multimodal} where yet again a rasterised image of the environment was utilised for the prediction of agents' future motion. However, in comparison to the previous study where a single motion for the agent of interest was anticipated, this work tackled the issue of reasoning about multi-modal nature of the task by forecasting number of candidate future trajectories along with the estimate of their probabilities. Vast majority of discussed methods focus on formulating the discussed task as regression task. Recently proposed model named \textit{CoverNet} \cite{phan2020covernet} proposes to frame this rather as a classification over a set of dynamically generated trajectories. The main motivation arises from an assumption that given the current state of an tracked agent within an environment, there exist a finite amount of relatively reasonable actions that can executed within a short period of prediction horizon.


\section{PROPOSED APPROACH}
\subsection{Problem Formulation and Notation}
We assume an availability of an appropriate tracking module that is capable of producing data corresponding to the state of a tracked agent at a fixed interval e.g. $2$Hz. First, let $\tau, \rho$ denote time-steps for the prediction and observed time horizon respectively. Next, let  $\mbox{\boldmath$ S$} = [\mbox{\boldmath$s$}_{t-\rho}, \cdots, \mbox{\boldmath$s$}_{t-1}, \mbox{\boldmath$s$}_{t}]$ represent a standardised matrix of the tracked agent's motion state from time-step $t-\rho$ to the initial time-step $t$ where $ \mbox{\boldmath$s$}_t = [v, a, \Delta \vartheta]$ denotes a vector of scalar features containing velocity, acceleration and heading change rate respectively. Moreover, let $\mbox{\boldmath$ P$} = [\mbox{\boldmath$p$}_{t-\rho}, \cdots, \mbox{\boldmath$p$}_{t-1}, \mbox{\boldmath$p$}_{t}]$ correspond to the agent's observed past positions, and $\mbox{\boldmath${Y}$} = [\mbox{\boldmath${y}$}_{t+1}, \mbox{\boldmath${y}$}_{t+2}, \cdots, \mbox{\boldmath${y}$}_{t+\tau}]$ to its future ground-truth positions with each vector in both matrices containing $(x, y)$ coordinates in agent's frame of reference. 

Additionally, we assume an access to HD maps that define the following road layers; road segments, drivable areas, lanes, walkways. Let the normalised matrix \mbox{\boldmath$M$} define a global chunk of a rasterised HD map which captures $100$ meters of the surrounding area in front of the vehicle as well as $5$ meters from the rear. The matrix \mbox{\boldmath$M$} is further extracted with accordance to the agent's position at the initial time-step $t$ and rotated towards its frame of reference. Furthermore, we define a normalised tensor $\mbox{\boldmath${\mathsf D}$} = [\mbox{\boldmath$\mathsf L$}_{t-\rho}, \cdots, \mbox{\boldmath$\mathsf L$}_{t-1}, \mbox{\boldmath$\mathsf L$}_{t}]$ which encapsulates local map chunks ($20m \times 20m$) with respect to the tracked agent (further details regarding local map chunks can be found in \cite{dulian2021exploiting}). Then, each $\mbox{\boldmath$\mathsf L$} \in \mbox{\boldmath$\mathsf D$}$ is unraveled into separate semantic layers such that $\mbox{\boldmath$\mathsf L$}_t = [\mbox{\boldmath$L$}_{t,0}, \mbox{\boldmath$\ L$}_{t,1}, \cdots, \mbox{\boldmath$ L$}_{t,n}]$ where $n$ is equal to number of road layers e.g. drivable areas, and $\mbox{\boldmath$ L$}_{t,i}$ is equal to a sparse matrix that encapsulates spatial data of semantic layer of type $i$. An example of both global and local HD map data is presented in Fig. \ref{fig:map_dis}.

Moreover, our goal is to train a generative model based on the CVAE framework which is generally composed as a three-part model with $Q_{\Theta}(\mbox{\boldmath$z$} \vert \mbox{\boldmath$x$}, \mbox{\boldmath$y$})$ (recognition network), $P_{\phi}(\mbox{\boldmath$z$} \vert \mbox{\boldmath$x$})$ (conditional prior) and $P_{\theta}(\mbox{\boldmath$y$} \vert \mbox{\boldmath$x$}, \mbox{\boldmath$z$})$ (generation network) parameterised by $\Theta, \phi$ and $\theta$ with each part of the model being commonly defined as a MLP. In $P_{\phi}(\mbox{\boldmath$z$} \vert \mbox{\boldmath$x$})$ the latent variable \mbox{\boldmath$z$} is conditioned on an arbitrary input \mbox{\boldmath$x$}, however, this can be further relaxed such that $P_{\phi}(\mbox{\boldmath$z$} \vert \mbox{\boldmath$x$}) = P_{\phi}(\mbox{\boldmath$z$})$ therefore making \mbox{\boldmath$z$} statistically independent \cite{kingma2014semi}. Finally, we are interested in using the trained generator to predict a diverse set of $k$ trajectories with respect to a tracked entity which we denote as $\mbox{\boldmath$\hat{\mathsf Y}$} = [\mbox{\boldmath$\hat{Y}$}_{0}, \mbox{\boldmath$\hat{Y}$}_{1}, \cdots, \mbox{\boldmath$\hat{Y}$}_{k}]$ with each matrix $\mbox{\boldmath$\hat{Y}$}_{i}$ being further constructed as $\mbox{\boldmath$\hat{Y}$}_{i} = [\mbox{\boldmath$\hat{y}$}_{i, t+1}, \mbox{\boldmath$\hat{y}$}_{i, t+2}, \cdots, \mbox{\boldmath$\hat{y}$}_{i, t+\tau}]$ where $\mbox{\boldmath$\hat{y}$}_{i, t+1}$ corresponds to a vector containing future coordinates of the trajectory $i$ at time $t+1$ such that $\mbox{\boldmath$\hat{y}$}_{i, t+1} = (\hat{x}_{i, t+1}, \hat{y}_{i, t+1})$ in the agent's frame of reference.

\subsection{State Encoding}
As previously discussed, recently presented approaches encode the rasterised HD map data using some form of previously proposed CNN architecture e.g. ResNet-50 (often pre-trained on a distinct domain) which lacks the ability to capture the equivariance with respect to detected features and discards useful information as a result of pooling operation. We propose to remedy these drawbacks with use of spatial encoder based on the CapsNet architecture presented in \cite{dulian2021exploiting}. We construct two separate feature extractors in order to encode both $\mbox{\boldmath${\mathsf D}$}$ and \mbox{\boldmath${M}$}.

We now demonstrate an example of encoding a single layer $\mbox{\boldmath${L}$}_{t,i} \in \mbox{\boldmath${\mathsf L}$}_t$ at time $t$ with the proposed layer encoder to extract the final latent representation vector of each semantic map \mbox{\boldmath${\mathsf L}$} over the observed time horizon. First of all, the matrix $\mbox{\boldmath${L}$}_{t,i}$ is passed through the convolutional base with \textit{Leaky ReLU} \cite{maas2013rectifier} non-linearity (we use the default value for negative slope i.e. $1e-2$):
\begin{equation}
    \mbox{\boldmath$\hat{\mathsf L}$}_{t,i} = \Phi_{\mathrm{l\_base}}(\mbox{\boldmath$L$}_{t,i})
    \label{eq:l_conv_base}
\end{equation}
which results in the tensor $\mbox{\boldmath$\hat{\mathsf L}$}_{t,i}$ which encapsulates activities of local low-level features. The extracted features are then passed through lower capsules:
\begin{equation}
    \mbox{\boldmath$\hat{L}$}_{t,i} = \Phi_{\mathrm{l\_lower}}(\mbox{\boldmath$\hat{\mathsf L}$}_{t,i})
    \label{eq:lower_caps}
\end{equation}
which results in the output matrix $\mbox{\boldmath$\hat{L}$}_{t,i}$ that contains $m$ capsules where each capsule is an $n-$dimensional vector with entries corresponding to instantiation parameters of detected features e.g. lines. In the former step we mentioned the use of scalar activation function i.e. Leaky ReLU, however, for all capsule layers we instead adopt \textit{squashing} non-linearity \cite{sabour2017dynamic} to normalize input vector \mbox{\boldmath$v$} as well as its magnitude to 0 or just below 1:
\begin{equation}
    \Phi_{\mathrm{squash}}(\mbox{\boldmath$v$}) = \frac{\|\mbox{\boldmath$v$}\|^2}{1+\|\mbox{\boldmath$v$}\|^2}\frac{\mbox{\boldmath$v$}}{\|\mbox{\boldmath$v$}\|}
    \label{eq:squash}
\end{equation}
Local low level features (e.g. lines, edges) between unraveled layers $[\mbox{\boldmath$L$}_{t,0}, \mbox{\boldmath$\ L$}_{t,1}, \cdots, \mbox{\boldmath$ L$}_{t,n}]$ of a semantic layer $\mbox{\boldmath${\mathsf L}$}_t$ resemble strong similarities and it is thus reasonable to share weights of former capsule layer to encode them. However, to account for the fact that their final representation demonstrates significant differences we decided to construct a $n$ number of higher capsules where $n$ is equal to the number of distinct type of road layers. Therefore, the output of the $\Phi_{\mathrm{l\_lower}}(\cdot)$ is further encoded with its respective higher capsule layer $i$:
\begin{equation}
    \mbox{\boldmath$\hat{l}$}_{t,i} = \Phi_{\mathrm{l\_higher\_i}}(\mbox{\boldmath$\hat{L}$}_{t,i})
    \label{eq:higher_caps}
\end{equation}
with vector $\mbox{\boldmath$\hat{l}$}_{t,i}$ encapsulating encoded parameters of the layer $\mbox{\boldmath${L}$}_{t,i}$. The above operation is further repeated for each $\mbox{\boldmath${L}$}_{t,i} \in \mbox{\boldmath${\mathsf L}$}_t$ with the final encoding of each layer being concatenated to create an input matrix for the final capsule:
\begin{equation}
    \mbox{\boldmath$\hat{l}$}_{t} = \Phi_{\mathrm{l\_final}}(\Phi_{\mathrm{concat}}(\mbox{\boldmath$\hat{l}$}_{t,0}, \mbox{\boldmath$\hat{l}$}_{t,1}, \cdots, \mbox{\boldmath$\hat{l}$}_{t,n}))
    \label{eq:final_caps}
\end{equation}
where $\mbox{\boldmath$\hat{l}$}_{t}$ encodes the final representation of $\mbox{\boldmath${\mathsf L}$}_t$. Repeating this process for each $\mbox{\boldmath${\mathsf L}$}_t \in \mbox{\boldmath${\mathsf D}$}$ yields matrix $\mbox{\boldmath$\hat{L}$}$ whose entries encapsulate encoding of all semantic layers from time $t-\rho$ to $t$ such that $\mbox{\boldmath$\hat{L}$} = [ \mbox{\boldmath$\hat{l}$}_{t-\rho}, \cdots, \mbox{\boldmath$\hat{l}$}_{t-1}, \mbox{\boldmath$\hat{l}$}_{t}]$.

Furthermore, in order to obtain the final encoding of the agent's state we concatenate the matrix $\mbox{\boldmath$\hat{L}$}$ with corresponding elements from motion state matrix \mbox{\boldmath$S$}:
\begin{equation}
    \mbox{\boldmath$\hat{s}$}_{t} = \Phi_{\mathrm{concat}}(\mbox{\boldmath$\hat{l}$}_{t}, \Phi_{\mathrm{state}}(\mbox{\boldmath$s$}_{t})), \quad \forall\ \mbox{\boldmath$\hat{l}$}_{t} \in \mbox{\boldmath$\hat{L}$}, \mbox{\boldmath$s$}_{t} \in \mbox{\boldmath$S$}
    \label{eq:z_s}
\end{equation}
to create encoded state matrix $\mbox{\boldmath$\hat{S}$} =[\mbox{\boldmath$\hat{s}$}_{t-\rho}, \cdots, \mbox{\boldmath$\hat{s}$}_{t-1}, \mbox{\boldmath$\hat{s}$}_{t}]$ where $\Phi_{\mathrm{state}}(\cdot)$ is a single fully-connected layer with Leaky ReLU. Lastly, the matrix $\mbox{\boldmath$\hat{S}$}$ is passed through a LSTM layer to compute final state representation in a temporal manner $\mbox{\boldmath$\hat{s}$} = \Phi_{\mathrm{state\_lstm}}(\mbox{\boldmath$\hat{S}$})$ with vector \mbox{\boldmath$\hat{s}$} corresponding to $\Phi_{\mathrm{state\_lstm}}(\cdot)$ final hidden-state.

\subsection{Recognition Network}
In addition to computing the encoding of the tracked agent's state \mbox{\boldmath$\hat{s}$} we also encode scene context \mbox{\boldmath${M}$} as well as agent's ground-truth future \mbox{\boldmath${Y}$} and past motion \mbox{\boldmath${P}$}. We obtain the encoding of the surrounding by running \mbox{\boldmath${M}$} through another CapsNet in a similar manner as we did with \mbox{\boldmath${\mathsf D}$}:
\begin{equation}
    \mbox{\boldmath$\hat{m}$} = \Phi_{\mathrm{m\_higher}}(\Phi_{\mathrm{m\_lower}}(\Phi_{\mathrm{m\_base}}(\mbox{\boldmath${M}$})))
    \label{eq:scene_caps}
\end{equation}
Next, both \mbox{\boldmath${Y}$} and \mbox{\boldmath${P}$} are flattened along their temporal dimensions to create vectors \mbox{\boldmath${g}$} (future) and \mbox{\boldmath${p}$} (past) of shape $2\times\tau$ and $2\times\rho$ respectively. We pass these vectors through their respective fully-connected layers with Leaky ReLU to get $\mbox{\boldmath${\hat{g}}$} = \Phi_{\mathrm{gt}}(\mbox{\boldmath${{g}}$})$ and $\mbox{\boldmath${\hat{p}}$} = \Phi_{\mathrm{past}}(\mbox{\boldmath${{p}}$})$ (we noticed no improvement when encoding \mbox{\boldmath${Y}$} and \mbox{\boldmath${P}$} through LSTM layers and therefore decided to use fully-connected based networks to account for computational efficiency).

We further define our recognition network in CVAE framework as $Q_{\Theta}(\mbox{\boldmath$z$} \vert \mbox{\boldmath$\hat{g}$}, \mbox{\boldmath$c$})$ where $Q_{\Theta}$ is constructed as a two-headed MLP each composed of two fully-connected layers with Leaky ReLU and batch normalisation \cite{ioffe2015batch}, and vector \mbox{\boldmath$c$} is the result of the following $\mbox{\boldmath$c$} =  \Phi_{\mathrm{concat}}(\mbox{\boldmath$\hat{p}$}, \mbox{\boldmath$\hat{m}$})$. $Q_{\Theta}$ aims to approximate the prior distribution $P_{\phi}(\mbox{\boldmath$z$})$ which we model as Gaussian such that $P_{\phi}(\mbox{\boldmath$z$}) = \mathcal{N}(\mbox{\boldmath$0$, \mbox{\boldmath$\mathbf{I} $}})$ by predicting distribution parameters i.e. \mbox{\boldmath$\hat{\mu}$} and \mbox{\boldmath$\hat{\sigma}$} of a latent variable \mbox{\boldmath$z$} which results in $\mbox{\boldmath$z$} \sim Q_{\Theta}(\mbox{\boldmath$z$} \vert \mbox{\boldmath$\hat{g}$}, \mbox{\boldmath$c$}) = \mathcal{N}(\mbox{\boldmath$\hat{\mu}$}, \mbox{\boldmath$\hat{\sigma}$})$ and leads to learning of the hidden representation of \mbox{\boldmath$\hat{g}$} given \mbox{\boldmath${c}$}.

\subsection{Motion Generator}
Our generation network which we refer to as motion generator is formulated as $P_{\theta}(\mbox{\boldmath$\hat{Y}$} \vert \mbox{\boldmath$z$}, \mbox{\boldmath$c$})$ and defined as an MLP with four fully-connected layers and Leaky ReLU non-linearity in-between layers. In addition to using $\mbox{\boldmath$z$}$ and $\mbox{\boldmath$c$}$ during the generation process, we also add the agent's encoded state $\mbox{\boldmath$\hat{s}$}$ as an additional intermediate input to the second fully-connected layer. The final layer of the generator yields a vector of size $2(x,y) \times \tau$ where $\tau$ corresponds to the future time horizon. The output of the generator is finally reshaped to create target matrix $\mbox{\boldmath$\hat{Y}$} = [\mbox{\boldmath$\hat{y}$}_{t+1}, \mbox{\boldmath$\hat{y}$}_{t+2}, \cdots, \mbox{\boldmath$\hat{y}$}_{t+\tau}]$ with each entry vector containing predicted future coordinates $(x,y)$ in agent's coordinate frame from time $t+1$ to $t+\tau$. During training phase we sample the latent variable $\mbox{\boldmath$z$}$ from $Q_{\Theta}$ i.e. $\mbox{\boldmath$z$} \sim Q_{\Theta}(\mbox{\boldmath$z$} \vert \mbox{\boldmath$\hat{g}$}, \mbox{\boldmath$c$})$, however, during testing the encoding of ground-truth motion $\mbox{\boldmath$\hat{g}$}$ is not available and we therefore sample $\mbox{\boldmath$z$}$ directly from the prior distribution $\mbox{\boldmath$z$} \sim P_{\phi}(\mbox{\boldmath$z$})$. Note that generating $k$ samples can be performed by sampling $\mbox{\boldmath$z$}$ from prior for $k$ times to create $\mbox{\boldmath${Z}$} = [\mbox{\boldmath${z}$}_{0}, \mbox{\boldmath${z}$}_{1}, \cdots, \mbox{\boldmath${z}$}_{k}]$ with $\mbox{\boldmath${z}$}_{i}$ corresponding to latent sample at index $i$, and then decoding the $i^{th}$ motion sequence such that:
\begin{equation}
    \mbox{\boldmath$\hat{Y}$}_i = P_{\theta}(\mbox{\boldmath$\hat{Y}$}_i \vert \mbox{\boldmath$z$}_i, \mbox{\boldmath$c$}) \quad \forall\ \mbox{\boldmath$z$}_i \in \mbox{\boldmath$Z$}
    \label{eq:z_k}
\end{equation}

Although the above explanation represents computation process of motion prediction for a single agent, it is trivial to extend the method for a multi-agent scenario. For the purpose of simplicity and to focus on presenting the main aspects of the paper we maintained a case of a single-agent.

\section{EXPERIMENTS AND RESULTS}

\subsection{Loss, Training Setup and Evaluation Metrics}
Note, we refer to our model as MMST (multi-modal stochastic trajectories) throughout the remainder of the paper. We focus on optimising the network by minimising the following loss function:
\begin{equation}
    \mathcal{J} = \alpha \mathcal{J}_1 + \beta \mathcal{J}_2
    \label{eq:loss}
\end{equation}
where $\mathcal{J}_1$ aims to minimise the distance between $Q_{\Theta}(\mbox{\boldmath$z$} \vert \mbox{\boldmath$\hat{g}$}, \mbox{\boldmath$c$})$ as well as the prior $P_{\phi}(\mbox{\boldmath$z$})$:
\begin{equation}
    \begin{split}
    \mathcal{J}_1 &= -D_{KL}(Q_{\Theta}(\mbox{\boldmath$z$} \vert \mbox{\boldmath$\hat{g}$}, \mbox{\boldmath$c$}) \vert \vert P_{\phi}(\mbox{\boldmath$z$}))\\
    &= \frac{1}{2}\sum_{i=1}^{n}(1+log((\mbox{\boldmath$\hat{\sigma}$}_i)^2) - (\mbox{\boldmath$\hat{\mu}$}_i)^2 - (\mbox{\boldmath$\hat{\sigma}$}_i)^2)
    \end{split}
    \label{eq:dkl}
\end{equation}
and $\mathcal{J}_2$ serves as reconstruction loss between $k$ predictions per sample and ground truth such that:
\begin{equation}
    \begin{split}
    \mathcal{J}_2 &= MoN(\mbox{\boldmath$Y$},\mbox{\boldmath$\hat{\mathsf Y}$})\\
                  &= \sum_{i=1}^{n} \min_{k} \mathrm{d}(\mbox{\boldmath$y$}_i, \mbox{\boldmath$\hat{y}$}_i^k)
    \label{eq:rec}    
    \end{split}
\end{equation}
where $\mathrm{d}(\cdot)$ refers to an arbitrary distance function such as $\mathrm{d}(\mbox{\boldmath$y$}_i, \mbox{\boldmath$\hat{y}$}_i^k) = \|\mbox{\boldmath$y$}_i - \mbox{\boldmath$\hat{y}$}_i^k\|^2$. We balance both loss terms by setting $\alpha = 1$ and $\beta = 1e-2$. Furthermore, we train the network for 360 epochs with \textit{SPS} optimizer with adaptive learning rate \cite{loizou2020stochastic} and with the batch size of $64$. Next, for every sample we observe 2 seconds of agent's past states i.e. $\rho \equiv 2$ seconds and we predict $k$ future samples each representing $\tau \equiv 6$ seconds of its future motion. MMST was implemented using PyTorch \cite{paszke2019pytorch} and trained/tested on a single Nvidia RTX 2080Ti. We further employ the following metrics to assess the quantitative performance of our model:
\begin{equation}
    \mathrm{minADE}_k = \frac{1}{n}\sqrt{\sum_{i=1}^{n}\min_{k}\| \mbox{\boldmath${y}$}_i -  \mbox{\boldmath$\hat{y}$}_i^k \| ^2} 
    \label{eq:minade}
\end{equation}
which measures minimum average displacement between all predictions of sample $i$ as well as minimum final displacement between sample's final prediction at time $t+\tau$:
\begin{equation}
    \mathrm{minFDE}_k = \frac{1}{n}\sqrt{\sum_{i=1}^{n}\min_{k}\| \mbox{\boldmath${y}$}_{i,t+\tau} -  \mbox{\boldmath$\hat{y}$}_{i,t+\tau}^k \| ^2} 
    \label{eq:minfde}
\end{equation}
\subsection{Dataset}
We report results of our experiments on the publicly available self-driving dataset \textit{nuTonomy Scenes} (nuScenes) \cite{caesar2020nuscenes} designed for a variety of tasks such as detection, tracking as well as motion prediction. nuScenes provides access to 1000 scenes (approximately 20 seconds each) that were collected in Boston and Singapore. Collected scenes provide a wide diversity with regards to weather conditions, traffic situations and traffic density. In addition, nuScenes contains human-annotated vectorized maps with 11 different semantic layers. Scenes and objects within each scene (e.g. vehicles, pedestrians) were accurately annotated at the rate of $2Hz$ and modeled as a cuboid, providing further access to object's position, size and yaw angle. Dataset is further split in accordance to nuScene's $train$ and $val$ split sets \footnote[1]{Available on the official repository}. Since the $test$ set has not been annotated we split the $train$ set into train and validation sets and use the $val$ split as a test set resulting in $20965/5000/5956$ split with regards to the final size of $train/val/test$ sets respectively.

\subsection{MoN - Optimal N training samples}\label{ssec:mon_optimal}
We first examine various $n$ values with regards to the training of the model with proposed MoN loss and the default distance function $d(\cdot)=L2$ where $n \in \{16, 32, 64, 128, 256\}$, results are presented in Table \ref{tb:mon_n}. We notice that for our proposed method the optimal number of samples during training is $\approx 32$ which yields least error across three out of four sampling scenarios ($k=10$). Moreover, the performance of the model does not increase with larger $n$ training samples (above $n=32$) but rather decreases. We also notice that for $k=10$ the best results are obtained when $n=16$ meaning that our method trained on MoN can perform optimally by being trained on relatively small sample size.

\begin{table}[h]
\caption{Comparison study between different settings of $n$ during training with respect to MoN loss. We provide minADE/minFDE error in meters for four different sampling values $k$.}
\label{tb:mon_n}
    \begin{center}
    \setlength{\extrarowheight}{3pt}
      \begin{tabular}{
        |l||
        >{\centering\arraybackslash}p{1.25cm}| >{\centering\arraybackslash}p{1.2cm}|
        >{\centering\arraybackslash}p{1.2cm}| >{\centering\arraybackslash}p{1.2cm}|
        >{\centering\arraybackslash}p{1.2cm}| 
        }
        \hline
        \multirow{2}{*}{\shortstack{MoN$_n$}} & 
        \multicolumn{4}{c|}{minADE$_k$/minFDE$_k$} \\
        \cline{2-5}
         & $k={10}$ & $k={25}$ & $k={50}$ & $k={100}$ \\
        \hline
        \hline
        MMST$_{16}$   & \textbf{1.77/3.81} & 1.32/2.51 & 1.09/1.87 & 0.92/1.38 \\
    
        MMST$_{32}$   & 1.82/3.92 & \textbf{1.25/2.42} & \textbf{1.05/1.82} & \textbf{0.89/1.32} \\
        
        MMST$_{64}$   & 1.99/4.37 & 1.43/2.79 & 1.14/2.00 & 0.95/1.46 \\
        
        MMST$_{128}$  & 2.07/4.53 & 1.45/2.84 & 1.17/2.02 & 0.94/1.44 \\
        
        MMST$_{256}$  & 2.19/4.80 & 1.52/3.00 & 1.19/2.11 & 0.97/1.50 \\
        
        \hline
        \end{tabular}
    \end{center}
\end{table}
\subsection{MoN - Distance function}
Furthermore, in Table \ref{tb:loss_mon} we present results of employing three different distance functions $d(\cdot)$ during training of the model (we set $n$ to the fixed value of $32$ as it presented most optimal performance during previous experiment). We examine popular $L1$ and $L2$ metrics as well as their combination which is further balanced by $\lambda$ parameter which we set to $0.5$. Results clearly demonstrate advantages of using $L2$ which outperforms other metrics on majority of scenarios. The balanced combination of $L2$ and $L1$ however does present best results on couple of cases and performs relatively similarly to $L2$ suggesting that appropriate adjustment of $\lambda$ parameter might lead to $L2$ being outperformed. 

\begin{table}[h]
\caption{Ablation study between different distance functions. Again, we provide results using four different sampling rates ($k$) during testing.}
\label{tb:loss_mon}
    \begin{center}
    \setlength{\extrarowheight}{3pt}
      \begin{tabular}{
        |l||
        >{\centering\arraybackslash}p{1.25cm}| >{\centering\arraybackslash}p{1.2cm}|
        >{\centering\arraybackslash}p{1.2cm}| >{\centering\arraybackslash}p{1.2cm}|
        >{\centering\arraybackslash}p{1.2cm}| 
        }
        \hline
        \multirow{2}{*}{\shortstack{Distance\\Function}} & 
        \multicolumn{4}{c|}{minADE$_k$/minFDE$_k$} \\
        \cline{2-5}
         & $k={10}$ & $k={25}$ & $k={50}$ & $k={100}$ \\
        \hline
        \hline
        $L1$   & 1.94/4.00 & 1.45/2.72 & 1.20/2.05 & 1.00/1.51 \\
    
        $L2$   & 1.82/3.92 & \textbf{1.25/2.42} & \textbf{1.05}/1.82 &\textbf{ 0.89/1.32} \\
        
        $\lambda(L1+L2)$   & \textbf{1.79/3.83} & 1.30/2.49 & 1.07/\textbf{1.81} & 0.90/1.36 \\

        \hline
        \end{tabular}
    \end{center}
\end{table}

\begin{table*}[t]
\caption{Comparison study of our final model vs two recently proposed methods from the literature. Apart from providing results using minADE/minFDE we also include the number of learnable parameters (millions) to demonstrate the difference in size of all examined methods. For CoverNet$^{\epsilon}$ we adjust number of modes that each variant is trained on with accordance to $\epsilon$ value.}
\label{tb:sota_comp}
    \begin{center}
    \setlength{\extrarowheight}{3pt}
      \begin{tabular}{
        ||l|
        l|| >{\centering\arraybackslash}p{1.25cm}|
        >{\centering\arraybackslash}p{1.25cm}| >{\centering\arraybackslash}p{1.25cm}|
        >{\centering\arraybackslash}p{1.25cm}| >{\centering\arraybackslash}p{1.25cm}|
        >{\centering\arraybackslash}p{1.25cm}| >{\centering\arraybackslash}p{1.25cm}|
        >{\centering\arraybackslash}p{1.5cm}||
        }
        \hline
        \multirow{2}{*}{\shortstack{Method}} & 
        \multirow{2}{*}{\shortstack{$n$-Modes}} & 
        \multicolumn{7}{c|}{minADE$_k$/minFDE$_k$} &%
        \multirow{2}{*}{\shortstack{\shortstack{\#Params}}} \\
        \cline{3-9}
         &  & $k=1$ & $k=5$ & $k=10$ & $k=20$ & $k=50$ & $k=100$ & $k=200$ & \\
        \hline
        \hline
        CoverNet$^8$        & 64 & 5.37/11.62 & 2.82/5.45 & 2.43/4.10 & 2.26/3.30 & 2.21/3.11 & - & - & 32.0m \\
    
        CoverNet$^4$       & 415 & 5.38/12.11 & 2.96/6.17 & 2.31/4.33 & 1.86/3.07 & 1.51/2.02 & 1.40/1.60& 1.34/1.39 &33.5m\\
        
        CoverNet$^2$       & 2206 & 5.48/12.30 & 2.98/6.30 & 2.32/4.60 & 1.84/3.26 & 1.38/2.02 & 1.17/1.46& 1.04/1.09  & 40.9m\\
        
        MTP       & 64 & \textbf{4.07/9.36} & \textbf{2.33/5.07} & 1.84/\textbf{3.70} & 1.45/2.80 & 1.07/1.85 & - & - & 38.5m\\
        
        MTP       & 128 & 4.31/9.89 & 2.90/6.43 & 2.39/5.18 & 1.88/3.82 & 1.23/2.04 & 0.93/\textbf{1.25}& - & 45.0m \\
        
        MTP       & 256 & 4.96/11.25 & 2.93/6.75 & 2.29/5.04 & 1.80/3.72 & 1.14/1.87 & 0.94/1.28 & 0.88/1.13 & 58.1m \\
        \hline
        MMST       & 32 & 6.34/15.22 & 2.44/5.51 & \textbf{1.82}/3.92 & \textbf{1.39/2.76} & \textbf{1.05/1.82} & \textbf{0.89}/1.32 & \textbf{0.78}/\textbf{0.98} & 7.4m \\

        \hline
        \end{tabular}
    \end{center}
\end{table*}

\subsection{Displacement errors for large k}
\begin{figure}[h]
  \centering
   \includegraphics[scale=.45]{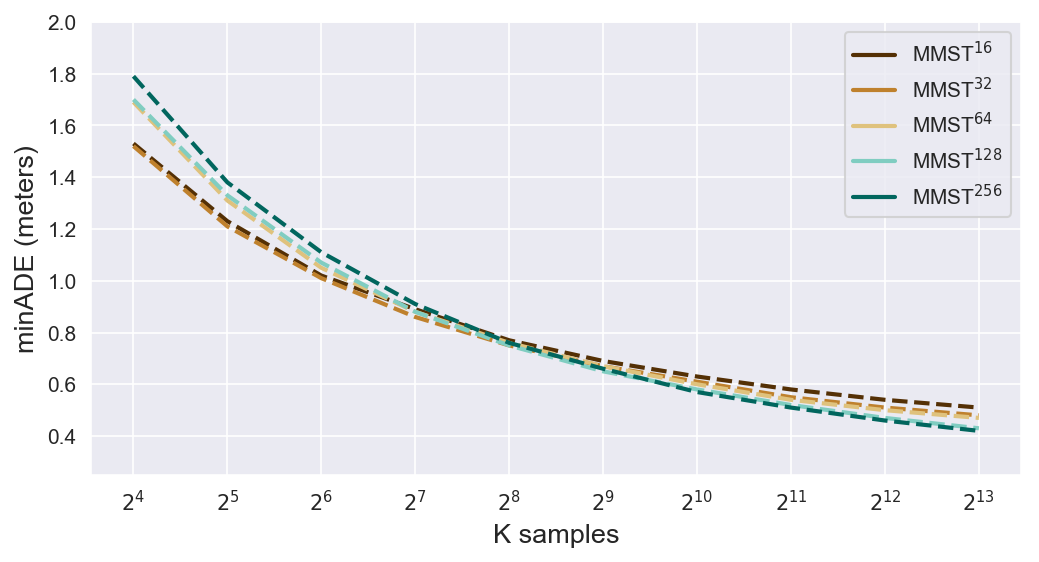}
  \caption{Results of the $\mathrm{minADE}$ error with regards to several MMST's variants where $ 2^4 \leq k \leq 2^{13}$.}
  \label{fig:ade}
\end{figure}

\begin{figure}[h]
  \centering
   \includegraphics[scale=.45]{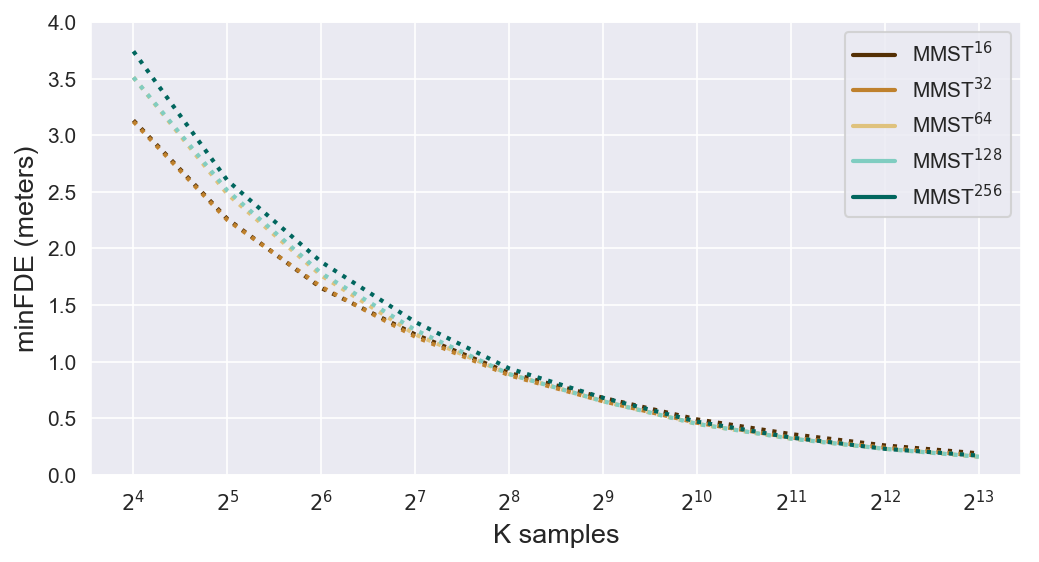}
  \caption{Results of the $\mathrm{minFDE}$ error with regards to several MMST's variants where $ 2^4 \leq k \leq 2^{13}$.}
  \label{fig:fde}
\end{figure}

Next, we analyse displacement errors (Fig. \ref{fig:ade} \& \ref{fig:fde}) with respect to a large number of sampled trajectories ($k\leq2^{13}$) using models presented in section \ref{ssec:mon_optimal}. As expected the displacement error for both metrics declines as the number of sampled trajectories grows. We notice that all variations of MMST perform similarly well with models trained on larger $n$ (i.e $128$ and $256$) achieving slightly lower $\mathrm{minADE}$ of about $\approx 0.40$.

Moreover, we see a similar trend with respect to the $\mathrm{minFDE}$ with all variations of MMST reaching roughly equivalently low error as $k$ grows. Interestingly, when $k>2^{9}$ the $\mathrm{minFDE}$ gets smaller than $\mathrm{minADE}$ with lowest displacement of roughly $\approx 0.17$.

\subsection{Comparison with recent work}
Finally, we present the performance comparison of our technique against recently proposed state-of-the-art methods, more specifically:
\begin{itemize}
    \item \textbf{CoverNet} \cite{phan2020covernet}: This method as previously discussed in section \ref{sec:related} proposes to frame the multi-modal probabilistic trajectory prediction from a classification point of view by pre-generating $n$ trajectories and then training the model to reduce the loss with respect to a classified sample that is closest to the actual ground-truth. We train the model by following settings outlined in the original paper across three different error tolerance settings i.e. $\epsilon \in \{2,4,8\}$ with pre-trained ResNet-50 to extract road features and encode context of the environment.
    \item \textbf{MTP} \cite{cui2019multimodal}: Another method that uses rasterised top-down view of the environment and encodes its salient features with use of ResNet-50. As with CoverNet we also adopt training settings for MTP with accordance to the details outlined within the original paper. We re-train the model to output $n \in \{64,128,256\}$ deterministic trajectories along with their probabilities and then sample $k$ most probable motions.
\end{itemize}
Results of the comparison study are presented in Table \ref{tb:sota_comp}. As demonstrated, our generative model either outperforms both approaches or produces results that are relatively similar to MTP. We noticed that our method yields significantly higher error when $k=1$ but then instantaneously reaches similar results when $k>1$. In addition, MMST tends to outperform other methods across most $k$ samples on $\mathrm{minADE}$ metric whilst reaching significantly lower errors when $k=200$. Furthermore, we observe that our model achieves relatively low error whilst at the same time significantly reducing number of parameters within the network which we attribute to the use of our backbone feature extractor based on Capsule Network. Both MTP and CoverNet employ ResNet-50 to encode context of the environment which we noticed leads to an early overfitting during re-training of these models. Moreover, the pre-trained ResNet-50 is trained on a distinct domain \cite{deng2009imagenet}, and it is therefore arguable whether this approach actually leads to an encoding of meaningful features from the rasterised HD map. It is important to note that our method is not restricted with respect to the number of samples it can generate as compared to both CoverNet and MTP which are limited to producing a deterministic output that must contain an equal amount of trajectories with regards to $n$ training modes.

\section{CONCLUSION}
In this work we introduced a novel approach towards short-term motion prediction in complex, urban environments. In essence, our work introduces the combination of stochastic model based on CVAE framework as well as spatial encoder based on CapsNet that can be used to replace a generally employed pre-trained CNN (e.g. ResNet-50) to extract salient features and encode the context of a rasterised surrounding. Experiments on public dataset (nuScene) demonstrate that our method either exceeds or preforms relatively similarly in comparison to recently proposed methods from the literature, whilst significantly reducing size of the network. We further analyse various variants of MoN loss to determine which distance function yields lowest error with respect to employed metrics. In addition, our method offers a way to generate an infinite number of diverse motion samples rather than being restricted to generating a fixed number of deterministic trajectories. This could potentially be used not solely to account for further analysis with respect to on-road safety but also to generate large number of diverse samples that could be used to e.g. train a model in a similar fashion to CoverNet. In the future we aim to extend this approach to account for social interactions amongst on-road participants in order to restrict our model to generate samples that are socially acceptable with respect to interactions between vehicles.

\addtolength{\textheight}{0cm}   

\bibliographystyle{plain}
\bibliography{root}
\end{document}